\documentclass{egpubl}
 
\WsPaper           

\usepackage[T1]{fontenc}
\usepackage{dfadobe}  

\usepackage{cite}  
\BibtexOrBiblatex
\electronicVersion
\PrintedOrElectronic
\ifpdf \usepackage[pdftex]{graphicx} \pdfcompresslevel=9
\else \usepackage[dvips]{graphicx} \fi

\usepackage{egweblnk} 

\usepackage{xcolor}
\usepackage{makecell} 

\usepackage{xspace}

\makeatletter
\DeclareRobustCommand\onedot{\futurelet\@let@token\@onedot}
\def\@onedot{\ifx\@let@token.\else.\null\fi\xspace}

\def\eg{\emph{e.g}\onedot} 
\def\ie{\emph{i.e}\onedot}

\def\etal{\emph{et al}\onedot}
\makeatother

\usepackage[ruled]{algorithm2e}
\usepackage{amsmath,amssymb} 
\usepackage{subcaption}
\usepackage{tikz}
\usetikzlibrary{positioning}
\usetikzlibrary{fit}
\usepackage{VMV2022}
\title[HandFlow]{HandFlow: Quantifying View-Dependent 3D Ambiguity in Two-Hand Reconstruction with Normalizing Flow}

\author[J. Wang et al.]
{\parbox{\textwidth}{\centering
J. Wang$^1$\orcid{0000-0001-9102-411X},
 D. Luvizon$^1$,\orcid{0000-0002-5055-500X},
 F. Mueller$^2$\orcid{0000-0003-2036-9238},
 F. Bernard$^3$,
 A. Kortylewski$^{1,5}$\orcid{0000-0002-9146-4403},
 D. Casas$^4$\orcid{0000-0002-3664-089X},
 and C. Theobalt$^1$\orcid{0000-0001-6104-6625}}
\\
{\parbox{\textwidth}{\centering $^1$MPI Informatics \& Saarland Informatics Campus, Germany\\
$^2$2Google Inc\\
$^3$University of Bonn, Germany\\
$^4$Universidad Rey Juan Carlos, Spain\\
$^5$University of Freiburg, Germany\\
}
}
}

\begin{document}

\input{figures/TeaserFigureVMV}

\maketitle
\begin{abstract}
Reconstructing two-hand interactions from a single image is a challenging problem due to ambiguities that stem from projective geometry and heavy occlusions.
Existing methods are designed to estimate only a single pose, despite the fact that there exist other valid reconstructions that fit the image evidence equally well.
In this paper we propose to address this issue by explicitly modeling the distribution of plausible reconstructions in a conditional normalizing flow framework.
This allows us to directly supervise the posterior distribution through a novel determinant magnitude regularization, which is key to varied 3D hand pose samples that project well into the input image.
We also demonstrate that metrics commonly used to assess reconstruction quality are insufficient to evaluate pose predictions under such severe ambiguity.
To address this, we release the first dataset with multiple plausible annotations per image called MultiHands.
The additional annotations enable us to evaluate the estimated distribution using the maximum mean discrepancy metric.
Through this, we demonstrate the quality of our probabilistic reconstruction and show that explicit ambiguity modeling is better-suited for this challenging problem.


\end{abstract}  

\section{Introduction}
\label{sec:intro}

Reconstructing two interacting hands in 3D is an actively researched topic, as it enables applications in various areas of vision and graphics, including augmented and virtual reality, robotics, or sign language translation.
While earlier methods leverage multi-camera setups~\cite{ballan_eccv2012, sridhar_iccv2013} or depth sensors~\cite{Taylor_siggraphasia2017, mueller_siggraph2019}, recent works focus on using monocular RGB cameras to enable potential applications in mobile or wearable settings.

However, hand pose estimation from monocular RGB images is a very challenging problem. 
Hand interactions lead to severe occlusions; and monocular color images exhibit an inherent depth and scale ambiguity. 
Existing methods~\cite{wang2020rgb2hands,moon2020interhand2,Zhang2021twohand} aim to \textit{deterministically} estimate the relative depth between the two hands directly. 
However, this is prone to error in heavily occluded situations due to the ill-posed nature of the problem. 
For example, a small error in hand scale or depth can cause a significant difference in touch points and hence semantics of the interaction.
As a result, most methods evaluate each hand pose independently using the root-relative pose error which discards important information regarding the positioning of the hands.

Given these extreme challenges, we take a different approach and propose to explicitly model the ambiguities instead (see Fig.~\ref{fig:teaser}).
Inspired by previous work on reconstruction of human body and face~\cite{Kolotouros_2021_ICCV, WehRud2021, kortylewski2018informed, schonborn2017markov}, we propose to predict a distribution over likely two-hand poses.
To this end, we adopt normalizing flow~\cite{rezende2015variational} as a way to parameterize the posterior distribution that enables not only fast sampling but also differentiable likelihood estimates.
This allows us to formulate a novel loss to supervise the shape of the distribution.
Our proposed regularization term encourages diversity in distribution without sacrificing image consistency, which is key to model the severe ambiguities in our setting. 

We quantitatively demonstrate that our sampled reconstructions capture the range of plausible articulations better than existing state-of-the-art methods.
This is facilitated by our new dataset, MultiHands, the first to provide multiple plausible annotations per image for measuring the accuracy of distribution predictions. 

In summary, our main contributions are: 
\begin{itemize}
    \item A method for reconstructing two-hand interactions that can generate diverse 3D poses which match the observed image.
    \item A new regularization term for training conditional normalizing flow to encourage diversity of samples.
    \item The first dataset to account for pose ambiguity by providing multiple pose annotations. 
\end{itemize}
Finally, we demonstrate that the estimated pose distribution can be leveraged for unambiguous view-point selection, a downstream application not possible with deterministic approaches. 

\section{Related Work}
\label{sec:rel_work}

The majority of existing works investigate the reconstruction of a single hand in free air or with a rigid object. These methods use input data that ranges from multi-camera setups~\cite{ballan_eccv2012,sridhar_iccv2013},
over depth sensors~\cite{sridhar_cvpr2015, Malik_2020_CVPR},
to monocular color images~\cite{mueller_cvpr2018, Boukhayma_2019_CVPR, Hasson_2020_CVPR}.
However, estimating hand poses during interaction with another hand is a significantly greater challenge due to, for example, occlusion and similarity between hands.
Thus we focus our discussion on methods tackling reconstruction of two interacting hands.

\noindent\textbf{Multiple Hands.}
Few existing methods reconstruct two interacting hands in 3D.
Oikonomidis \etal~\cite{oikonomidis2012tracking} first leveraged a multi-camera setup to mitigate some of the inherent challenges like strong occlusions.
While some recent methods still use multi-view setups and markers~\cite{simon2017hand, han2018online}, most works have moved to employing single depth sensors for the flexibility of the capture setup~\cite{Taylor_siggraphasia2017, mueller_siggraph2019}.

Considering a monocular color image as input, 
Wang \etal~\cite{wang2020rgb2hands} and Moon \etal~\cite{moon2020interhand2} simultaneously developed the first two methods for 3D pose estimation of interacting hands.
While the former combines machine-learning-based pixel-to-model correspondence prediction with optimization-based model fitting, the latter uses a neural network to predict 3D joint positions directly.
Recent extensions make use of visibility~\cite{Kim_2021_ICCV} or hand part segmentation~\cite{fan2021digit} to help the network to take into account occlusion information. 
Others were developed to additionally estimate hand surfaces, either as a parametric model~\cite{Zhang2021twohand} or as a mesh \cite{Li2022intaghand}.
These methods have in common that they deterministically reconstruct the hand interaction. 
However, during interactions, we often observe heavy occlusions between hands or ambiguous semantics (see Fig.~\ref{fig:teaser}).
In contrast, our work focuses on tackling the specific challenges of such interactions through a probabilistic approach. 

\noindent\textbf{Probabilistic Methods.}
Explicitly accounting for ambiguities in monocular RGB images is an important problem that has received recent attention in the field of 3D human poses estimation and face reconstruction 
\cite{zanfir2020weakly,Kolotouros_2021_ICCV, WehRud2021, kortylewski2018informed, schonborn2017markov}.
However, the only existing method for hands is designed for a single hand and uses depth images as input~\cite{Ye_2018_ECCV}. 
Our approach is the first to address the more ambiguous case of RGB images with challenging two-hand interactions.

Some recent probabilistic solutions for estimating 3D pose of a single body~\cite{Kolotouros_2021_ICCV, WehRud2021} are also based on normalizing flows~\cite{rezende2015variational}, which can construct a complex distribution from a simple probability density using invertible operations. 
In our approach, we build upon normalizing flows to address severe monocular depth ambiguities and occlusions in two interacting hands scenarios.
To this end, we introduce a new regularization term for the pose distribution and show that it is crucial for encouraging the sample diversity needed to model the ambiguities.

\noindent\textbf{Two-Hand Datasets.}
Although several datasets exist that contain images of synthetic~\cite{mueller_siggraph2019, wang2020rgb2hands, lin2021two} or real \cite{tzionas_ijcv2016, simon2017hand, wang2020rgb2hands, moon2020interhand2} images of two interacting hands,
they all provide only a single annotation per image. 
This is insufficient for monocular RGB reconstruction since multiple plausible poses can fit the image equally well. 
We argue that these plausible poses should also be considered correct, and propose our new MultiHands dataset to extend the existing InterHand2.6M~\cite{moon2020interhand2} with $100$ additional annotations per image. 
With MultiHands, we are able to quantify the pose ambiguity in an image, and to use a new metric for measuring the distance between predicted and ground-truth pose distributions. 
\begin{figure*}
\includegraphics[width=\linewidth, trim={0 11.4cm 2.35cm 0},clip]{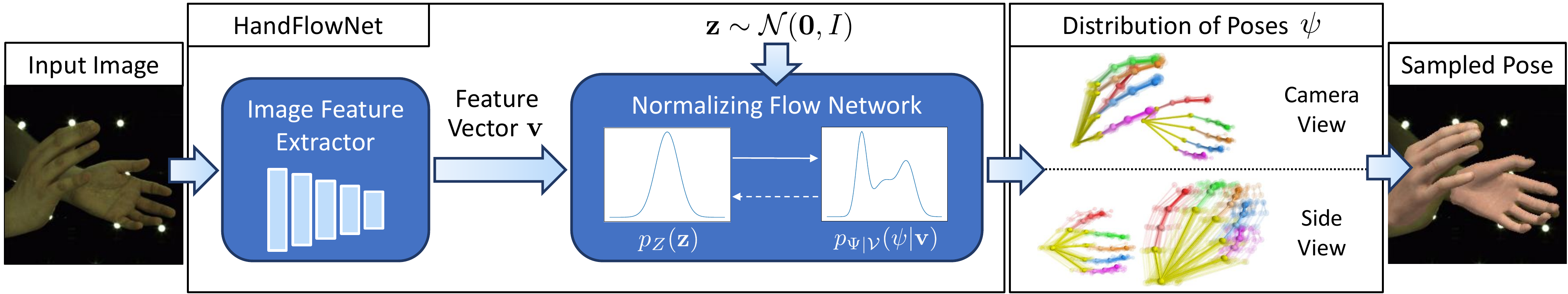}
\caption{
Our \emph{HandFlowNet} first extracts an image feature vector $\textbf{v}$ from 2D cues in the input image. 
The feature vector is then used as conditioning input to a normalizing flow network to output a distribution of 3D hand poses that plausibly explain the monocular input.
} 
\label{fig:overview}
\end{figure*}

\section{Method}
The goal of our method is to estimate a distribution of 3D hand poses that are plausible to explain a given monocular color image.
To this end, we propose \emph{HandFlowNet}. 
Our method first extracts a feature encoding from the input image, which we then use to generate the desired output pose distribution
from a normalizing flow network (Section~\ref{sec:handflownet}).
The estimated 3D hand poses are parameterized using the MANO hand model (Section~\ref{sec:mano_model}).

\subsection{Hand Model}
\label{sec:mano_model}

We use the MANO hand model~\cite{Romero_siggraphasia2017} to represent the hand surface as well as additional parameters for the rigid transformation. 
We will first describe the parameterization of a single hand, which is then readily expanded to two hands.

Given $15$ joint rotations $R \in \mathbb{R}^{15\times3\times3}$ represented as stacked rotation matrices and shape parameters $\beta \in \mathbb{R}^{10}$, the MANO model computes the hand surface as mesh and the 3D hand keypoint positions.
In order to place the hand correctly relative to the camera, we additionally estimate global rotation parameters $r \in \mathbb{R}^{3\times3}$, the hand root position in image coordinates $t \in \mathbb{R}^{2}$, and the perspective scale factor $s \in \mathbb{R}$.
This enables the recovery of the global pose when the focal length is known at inference ~\cite{Boukhayma_2019_CVPR}.
The combined global and joint rotations $\{r, R\}$ are parameterized using the 6 DOF representation $\theta \in \mathbb{R}^{16\times3\times2}$ as proposed in~\cite{Zhou_2019_CVPR} 
Therefore, the full set of parameters for a single hand is defined as $\psi = \{\theta, \beta, t, s\} \in \Psi$, where $\Psi$ denotes the parameter space,
and the full set of parameters for both hands is defined as $\psi_{\text{both}} = [\psi_{\text{right}}, \psi_{\text{left}}]$.
In the following, we will refer to $\psi_{\text{both}}$ simply as $\psi$.

\subsection{HandFlowNet}
\label{sec:handflownet}

Given a monocular input image, our \emph{HandFlowNet} regresses a distribution of 3D hand poses corresponding to plausible hand poses that could be observed in the image (see Fig.~\ref{fig:overview}).
\emph{HandFlowNet} can be divided into two parts, an image feature extractor and a conditional normalizing flow network that produces a 3D pose distribution and is conditioned on the extracted image feature vector.

\subsubsection{Image Feature Extractor}
\label{sec:featextractor}
The image feature extractor summarizes the visible, unambiguous features that the sampled poses should reconstruct.
We use ResNet-50~\cite{he_cvpr2016} as the backbone architecture. 
From an input image with resolution $224 \times 224$, we extract the $2048$-dimensional feature vector $\mathbf{v} \in \mathcal{V}$ from the average pooling of the last residual block, and use it as the conditional vector for the next step.

\subsubsection{Normalizing Flow Network}
\label{sec:normflownet}

To predict a range of plausible poses, we must first choose a way to parameterize a distribution $p_Y(\mathbf{y})$. 

Normalizing flow~\cite{rezende2015variational} does this by learning an invertible transformation $f: \mathbb{R}^d \rightarrow \mathbb{R}^d$ of a simple distribution $p_Z(\mathbf{z})$, \ie
\begin{equation}
    p_Y(\mathbf{y}) = p_Z(\mathbf{z}) \left\vert \operatorname{det} \frac{\partial f(\mathbf{z})}{\partial \mathbf{z}} \right\vert^{-1} \: .
\end{equation}
where $\mathbf{y} = f(\mathbf{z})$.
This invertible parameterization allows for both differentiable sampling and likelihood estimation.
As a result, we can apply losses on each sample to improve reconstruction quality, while supervising the entire distribution using negative log likelihood loss and multiple annotations (see Section~\ref{sec:losses}).

Since we want to estimate a distribution over the space of 3D hand poses $\Psi$ given an image feature vector $\mathbf{v}$,
we are interested in finding the conditional distribution $p_{\Psi|\mathcal{V}}(\psi|\mathbf{v})$. 
To this end, normalizing flow can be extended to conditional normalizing flow~\cite{winkler2019learning} by using transformations $f_\mathbf{v}: \mathbb{R}^d  \rightarrow \mathbb{R}^d$ parameterized by $\mathbf{v}$, so that
we have 
\begin{equation}
p_{\Psi|\mathcal{V}}(\psi|\mathbf{v}) = p_{Z|\mathcal{V}}(\mathbf{z}|\mathbf{v}) \left\vert\operatorname{det}\frac{\partial f_\mathbf{v}(\mathbf{z})}{\partial \mathbf{z}}\right\vert^{-1}    \: .
\end{equation}
For our implementation, we use the conditional GLOW architecture for $f_\mathbf{v}$ which has been successfully used in previous work~\cite{Kolotouros_2021_ICCV} due to its quick sampling and probability estimation.
For a more detailed overview of different architectures, we refer to~\cite{Kobyzev2021}. 

By setting $p_{Z|\mathcal{V}} = p_Z \sim \mathcal{N}(\mathbf{0},I)$, the mode of the distribution $p_{\Psi|\mathcal{V}}(\psi|\mathbf{v})$ can be obtained as $f_\mathbf{v}(\mathbf{0})$. 
We choose this design to provide easy access to the mode sample for use in our losses.

\subsubsection{Training Losses}
\label{sec:losses}

In the following, we detail the losses used for training. The entire loss is given by
\begin{equation}
\mathcal{L} =~
\mathcal{L}_{\text{nll}} + 
\mathcal{L}_{\text{DetMag}} +
\mathcal{L}_{\psi} + \\
\mathcal{L}_{\mathcal{J}_\text{3D}} + 
\mathcal{L}_{\mathcal{J}_\text{2D}} +
\mathcal{L}_{\theta} \,.
\label{eq:Total}
\end{equation}
Here, $\mathcal{L}_{\text{nll}}$ and $\mathcal{L}_{\text{DetMag}}$ are used to supervise the likelihood of the annotations, and
$\mathcal{L}_{\psi}$, $\mathcal{L}_{\mathcal{J}_\text{3D}}$,
$\mathcal{L}_{\mathcal{J}_\text{2D}}$,
and $\mathcal{L}_{\theta}$ are used to
supervise the quality of the sampled reconstructions.
For network training parameters and loss weights, please refer to the supplemental document.

\noindent\textbf{Maximum Likelihood Estimation.}
Given images and their 3D annotation, we want to ensure that the probability of the pose annotation $\psi^*$ is maximized. 
Hence, we minimize the negative log likelihood (NLL) loss
\begin{align}
\mathcal{L}_{\text{nll}} &= - \operatorname{ln} p_{\Psi|\mathcal{V}}(\psi^*|\mathbf{v}) \nonumber \\
&= - \operatorname{ln} p_{Z}(f_\mathbf{v}^{-1}(\psi^*)) \left\vert\operatorname{det}\frac{\partial f_\mathbf{v}(\mathbf{z})}{\partial \mathbf{z}}\right\vert^{-1} \: .
\label{eq:NLL}
\end{align}
When multiple annotations $\{..., \psi^*_{n}\}$ are available, the NLL loss is minimized over all annotated poses.

\noindent\textbf{Enhancing Pose Variety.}
We observe that training the network using just the term $\mathcal{L}_{\text{nll}}$ quickly collapses the variety in the output pose distribution.
To explain this, we note that $\mathcal{L}_{\text{nll}}$ maximizes $\left\vert\operatorname{det}\frac{\partial f_\mathbf{v}(\mathbf{z})}{\partial \mathbf{z}}\right\vert^{-1}$, which describes the compression factor between the two spaces for density conservation.
Therefore, the network can trivially optimize the conditional distribution by concentrating the density in the pose space, leading to the collapse in $p_{\Psi|\mathcal{V}}(\psi|\mathbf{v})$.
To prevent this, we add the regularization term
\begin{equation}
\mathcal{L}_{\text{DetMag}} = - \operatorname{ln}
\left\vert\operatorname{det}\frac{\partial f_\mathbf{v}(\mathbf{z})}{\partial \mathbf{z}}\right\vert \:.
\end{equation}
Since this term aims at increasing variation in the output distribution of the normalizing flow network only, we do not backpropagate it into the image feature extractor.
Otherwise, the extraction network might be hindered in learning pose-relevant features.

\noindent\textbf{Mode Supervision.}
While $\mathcal{L}_{\text{nll}}$ encourages the probability of the pose annotations to be maximized, we also want the mode sample $f_\mathbf{v}(\mathbf{0})$ to be a valid reconstruction.
We use the loss
\begin{equation}
\mathcal{L}_{\psi} = || f_\mathbf{v}(\mathbf{0}) - \psi^* ||^2_2\:.
\label{eq:loss_param}
\end{equation}
Note that $\mathcal{L}_{\psi}$ is complementary to $\mathcal{L}_{\text{nll}}$ and both together form a two-sided loss that ensures plausible pose predictions.
When multiple annotations are available, a single annotation is randomly chosen to act as the mode sample for the entire training procedure.

Although data with MANO parameter annotation exists, the amount is limited compared to the amount of data with joint position annotations.
To make use of all available data, we impose the additional 3D joint position loss
\begin{equation}
    \mathcal{L}_{\mathcal{J}_\text{3D}} = \sum_{i = 1}^{N_J} || \mathcal{J}(\psi)_i - P^{\text{3D}}_i ||^2_2 \: ,
\end{equation}
where $\mathcal{J}$ is a function defined by the hand model that calculates the 3D joint positions given pose parameters $\psi$, and $P^{\text{3D}}$ are the 3D joint position annotations.

\noindent\textbf{2D Consistency.}
Our \emph{HandFlowNet} aims to provide a distribution of poses that all correspond to the same input image.
Hence, the 2D position of visible joints should be the same for the mode and the samples of the distribution, and should thus match the annotation. We employ
\begin{equation}
\mathcal{L}_{\mathcal{J}_\text{2D}} = \sum_{i = 1}^{N_J} \eta_i \,|| \Pi(\mathcal{J}(\psi)_i) - P^{\text{2D}}_i ||^2_2 \: ,
\label{eq:loss_joints2D}
\end{equation}
where $\Pi$ is the known camera projection, $P^{\text2D}$ are the 2D joint position annotations, and $\eta_i = 1$ if the joint $i$ is visible and $0$ otherwise.
These visibility scores are computed from the meshes of MANO pose annotations. 
We calculate $\mathcal{L}_{\mathcal{J}_\text{2D}}$ on the mode of the distribution $f_\mathbf{v}(\mathbf{0})$ and on two samples from the estimated distribution $p_{\Psi|\mathcal{V}}(\psi|\mathbf{v})$.

\noindent\textbf{Rotation Regularization.}
As explained in Section~\ref{sec:mano_model}, we use the continuous 6-dimensional representation for 3D rotations proposed by Zhou \etal~\cite{Zhou_2019_CVPR}.
The representation is not unique, \ie, there are multiple $A \in \mathbb{R}^{3 \times 2}$ that represent the same 3D rotation $R \in \operatorname{SO}(3)$.
To encourage consistent output, we follow previous work~\cite{Kolotouros_2021_ICCV} and add a regularizer that constrains all rotations in their 6-dimensional representation $A$ to be orthonormal
\begin{equation}
    \mathcal{L}_{\theta} = \sum_{A \in \theta} ||A^{\top}A - I||^2_F \: .
\end{equation}
%

\section{Creating Additional Annotations}

While there exists a single \emph{ground-truth} pose, i.e.~the one that forms a given image, recovering this exact pose from an image is ambiguous since there are 
multiple \emph{plausible pose annotations}.
Since our goal is to model this ambiguity with a distribution, the single ground truth found in most datasets is not sufficient for evaluating our predictions and more annotations are needed.

Here we describe how we obtain additional annotations from a provided MANO ground truth. 

\noindent\textbf{Plausible Pose Annotations:}
Given the ground-truth pose parameters $\psi_{gt}$ and a camera projection $\Pi$, 
an annotation $\psi_{annot}$ is plausible if the hand joints fit the observed image and the overall articulation is anatomically possible. To ensure this, we use the following criteria: 
\begin{itemize}
    \item The 2D locations of visible joints should be within a pixel threshold of the ground truth locations.
    \item Occluded joints in the original pose should remain occluded.
    \item The pose should be anatomically likely as measured using the pose PCA space of the MANO model \cite{Romero_siggraphasia2017}.
    A likelihood threshold is used to eliminate extreme articulations.
    \item No collision between hands. 
    Collisions are detected using Gaussian proxies \cite{mueller_siggraph2019} attached to the MANO model. Collision occurs when the one-standard-deviation spheres of the Gaussian proxies intersect each other. 
\end{itemize}

\noindent\textbf{Annotation Generation:}
Starting from the ground-truth pose parameters $\psi_{gt}$, we perturb the hand pose parameters to generate hand pose proposals.
These proposals are checked for plausibility as defined in the above criteria, and implausible pose annotations are rejected. 
The accepted plausible pose annotations will now serve as new starting poses for the next iteration. 
This perturbation and plausibility checking is repeated for a fixed number of iterations to obtain the final plausible pose annotations. 
For additional implementation details, please refer to the supplemental document.

\input{figures/GroundTruthVisFigure}

\label{sec:annot_gen}
\section{Experimental Results}

We evaluate our method on existing datasets (Sec. \ref{sec:dataset}), and discuss the limitations of commonly used metrics in dealing with ambiguity (Sec. \ref{sec:alignment} and \ref{sec:problem_with_metric}). 
To deal with this ambiguity, we propose to use an alternative metric (Sec. \ref{sec:problem_with_metric}) to evaluate our method (Sec. \ref{sec:qualitative}, \ref{sec:ablation}, \ref{sec:qualitative}). 
Finally, we show an application beyond pose estimation to demonstrate the advantages of distribution estimation (Sec. \ref{sec:view_selection}). 

\input{figures/SoTAComparisonFigureNew}

\subsection{Datasets}
\label{sec:dataset}
Here we describe each dataset and practical considerations that we took into account to run the experiments.

\noindent\textbf{InterHand2.6M Dataset
\cite{moon2020interhand2}.}
We use the $673{,}514$ training frames labeled as interacting hands to train our method. 
Notice that the terms 
$\mathcal{L}_{\text{nll}}$,
$\mathcal{L}_{\psi}$ from Eq.~\ref{eq:NLL}, \ref{eq:loss_param}, respectively, require MANO parameter annotations. 
These losses are applied to the subset of $394{,}599$ frames where these are available.

Following the method of InterNet~\cite{moon2020interhand2}, we use RootNet~\cite{Moon_2019_ICCV_3DMPPE} for hand detection. A $334{\times}334$ crop centered around the provided bounding box is used for the image feature extractor.

\noindent\textbf{MultiHands Dataset:}
Using the method described in Sec.~\ref{sec:annot_gen}, we propose to extend InterHand2.6M with $100$ additional annotations for each of the $281,369$ test and $394,599$ training images with MANO annotations. 
Since our losses $\mathcal{L}_{\text{nll}}$ and $\mathcal{L}_{\text{DetMag}}$ can use multiple annotations, we also use MultiHands for training. See Fig.~\ref{fig:gt_vis} and the supplemental document for examples of annotations.

\noindent\textbf{Tzionas Dataset ~\cite{tzionas_ijcv2016}.}
To demonstrate that the learned 3D pose distribution generalizes to other settings,
we show qualitative results on the Tzionas Dataset.
This dataset has seven sequences captured in an office environment with only 2D annotations.

Following Moon \etal~\cite{moon2020interhand2}, we trained with mixed batches on $90\%$ of the annotated 2D frames, and show results on the remaining $10\%$.

\subsection{Pose Alignment}
\label{sec:alignment}
We use three different alignments to evaluate the mean per-joint position error (MPJPE) in mm.
All equations can be found in the supplemental document.

\noindent\textbf{Root-Relative MPJPE (RR)} captures the errors in articulation, where each hand is individually root-aligned. 
\noindent\textbf{Right-Root-Relative MPJPE (RRR)}
measures the accuracy of the two hands together, where both hands are aligned to just the root of the right hand.
\noindent\textbf{Global MPJPE (Global)}
captures the accuracy of the global pose estimate, \textit{without any alignment}. 

Although the RR metric is most commonly reported in the literature, it evaluates the two hands independently by ignoring the relative hand placements.
Since this placement is vital for most applications, we show and focus on the RRR and Global metrics.

\subsection{Problem with Traditional Metric}
\label{sec:problem_with_metric}
\begin{table}[t]
\begin{center}
\resizebox{0.48\textwidth}{!}
{
\begin{tabular}{|l|c|c|c|}
\hline
Method &  Global MPJPE$\downarrow$ & RRR MPJPE$\downarrow$  &  RR*  MPJPE $\downarrow$\\
\hline\hline
InterNet (min)& 67.2 & 24.5 & 22.6 \\ 
InterNet (max)& 103.6 & 42.2 & 24.6 \\
Fan et al. (min)& 65.7 & 27.1 & 20.5 \\ 
Fan et al. (max)& 102.1 & 45.9 & 22.5 \\
\hline
\end{tabular}   
}
\end{center} 

\caption{
MPJPE of deterministic estimates can vary wildly depending on the plausible pose annotation used. For RR*, the error is reported for occluded joints. All errors are in mm.
}
\vspace{-1mm}
\label{table:deterministic_error}
\end{table}

When the observed image is ambiguous,
the choice of the target pose can greatly impact the MPJPE even though equally valid alternative exist.
To quantify this effect on InterHand2.6M, we evaluated the InterNet~\cite{moon2020interhand2} and Fan et al.~\cite{fan2021digit} predictions against the closest and farthest annotation in MultiHands ( Table~\ref{table:deterministic_error}).

For the challenging Global and RRR metrics, the choice of plausible annotation accounts for a difference of $36$mm and $18$mm on average.
Even when each hand is evaluated independently with the RR metric, the occluded joints differ by $2$mm on average.

We argue that this sensitivity to the choice of annotation makes MPJPE unsuitable for the highly ambiguous monocular two-hand reconstruction task.
Instead, a metric that measures the distances between pose distributions would better reflect prediction quality.

\vspace{5pt}
\subsection{Maximum Mean Discrepancy (MMD)}

We can measure how well the estimated distribution matches the annotation distribution using the maximum mean discrepancy (MMD) \cite{gretton12aJMLR}.

The empirical MMD can be estimated given sampled pose predictions, multiple pose annotations, and the selection of a kernel function. 
We used 100 samples and annotations, and choose Gaussian kernels for our evaluation. 
All reported MMD are averaged over different kernel distance scales.

\subsection{Comparison to the State of the Art}
\label{sec:SoTA}
\begin{table}[t]
\begin{center}
\resizebox{0.48\textwidth}{!}
{
\begin{tabular}{|l|c|c|c|}
\hline
Method &  Global MMD $\downarrow$ & RRR MMD $\downarrow$  & RR MMD $\downarrow$ \\
\hline\hline
Ours & \textbf{0.50} & \textbf{0.42} & \textbf{0.44}\\
\hline
VAE & 0.61 & 0.47 & 0.48\\
Gaussian & 0.82 & 0.51 & 0.46\\
MCDropout & 0.91 & 0.60 & 0.51\\
\hline
InterNet & 1.12 & 0.59 & 0.56\\
Fan et al. & 1.12 & 0.63 & 0.50\\
\hline
\end{tabular}   
}
\end{center} 

\caption{
Our method best captures the true distribution of plausible poses using the maximum mean discrepancy (MMD)\cite{gretton12aJMLR}. 
}
\label{table:comparisonSoTA}
\end{table}

\noindent\textbf{Competing methods.}
We implement the widely applied probabilistic baselines Monte Carlo dropout (MC-dropout)~\cite{gal2016dropout}, aleatoric uncertainty (Gaussian)~\cite{kendallNIPS2017}, and Variational Auto Encoder (VAE)~\cite{kingma2014auto} for comparisons.
The implementation details can be found in the supplemental document.
As reference, we also compare against deterministic methods~\cite{moon2020interhand2, fan2021digit} by treating the estimates as a Dirac delta distribution.
Given each method, $100$ poses are sampled to find the MMD to ground-truth samples. 
MMD is computed for all alignment to better understand the sources of ambiguity. 

\noindent\textbf{Results.} 
Overall, our method produces estimates that best match the ground-truth distribution (Table \ref{table:comparisonSoTA}).
This is especially notable for the challenging Global and RRR MMD metric, which demonstrates the benefits of our formulation under ambiguity.
State-of-the-art deterministic methods fail to account for ground truth variability (Fig. \ref{figure:comparisonSTAR}).
As a result, they have one of the worst MMD.

For reference, a comprehensive evaluation of our method using the single provided InterHand2.6M annotation can be found in the supplemental document. 
There, we show that our best sample out-performs the state-of-the-art methods while still remaining competitive as a single pose estimator.

\begin{figure*}
  \centering
  \begin{subfigure}{0.33\columnwidth}
  \begin{tikzpicture}
  \hspace{-0.07cm}\node (image) at (0,0) {\includegraphics[trim=0 1.2cm 0 1.2cm, clip, width=\linewidth]{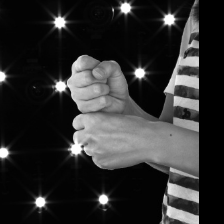}};
  \node[above = 1cm] {\footnotesize{Input}};
  \end{tikzpicture}
  \end{subfigure}
  \begin{subfigure}{0.33\columnwidth}
  \begin{tikzpicture}
  \hspace{-0.1cm}\node (image) at (0,0) {\includegraphics[trim=2cm 1cm 2cm 1cm, clip, width=\linewidth]{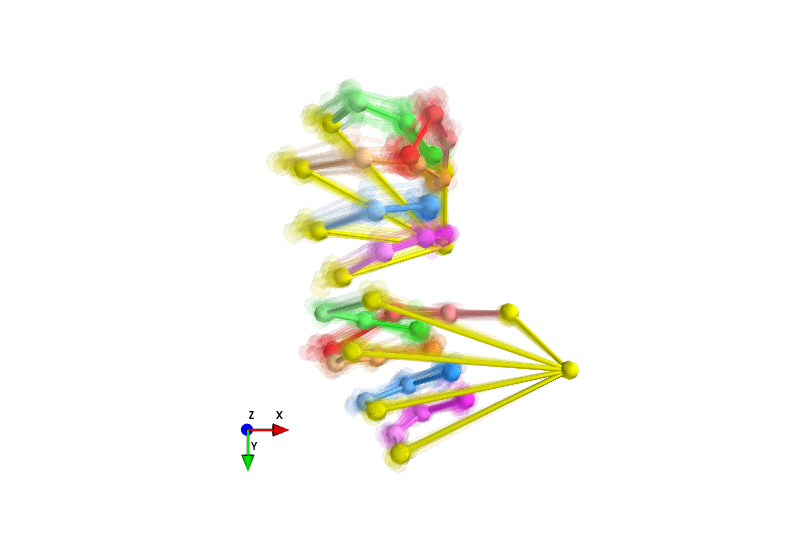}};
  \node[above = 1cm] {\footnotesize{Camera View}};
  \end{tikzpicture}
  \end{subfigure}
  \begin{subfigure}{0.33\columnwidth}
  \begin{tikzpicture}
  \hspace{-0.1cm}\node (image) at (0,0) {\includegraphics[trim=2cm 1cm 2cm 1cm, clip, width=\linewidth]{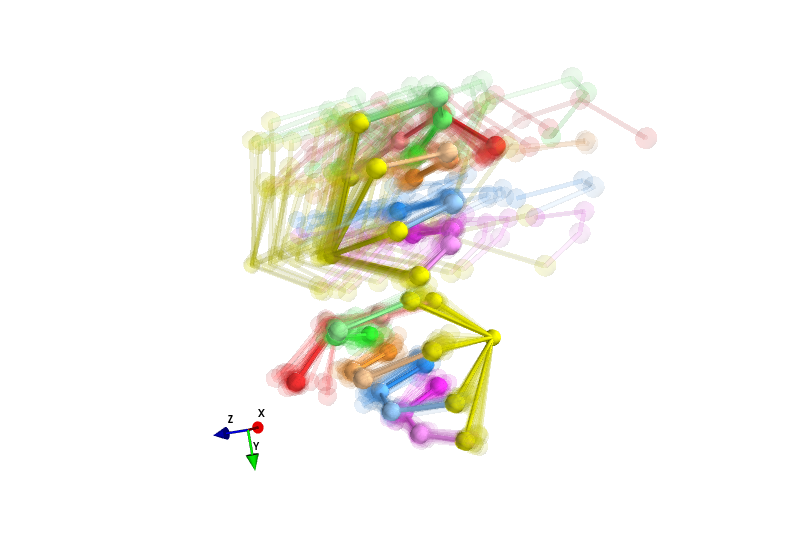}};
  \node[above = 1cm] {\footnotesize{Side View}};
  \end{tikzpicture}
  \end{subfigure}
  \begin{subfigure}{0.33\columnwidth}
  \begin{tikzpicture}
  \hspace{-0.07cm}\node (image) at (0,0) {\includegraphics[trim=0 1.2cm 0 1.2cm, clip, width=\linewidth]{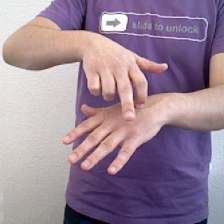}};
  \node[above = 1cm] {\footnotesize{Input}};
  \end{tikzpicture}
  \end{subfigure}
  \begin{subfigure}{0.33\columnwidth}
  \begin{tikzpicture}
  \hspace{-0.1cm}\node (image) at (0,0) {\includegraphics[width=\linewidth]{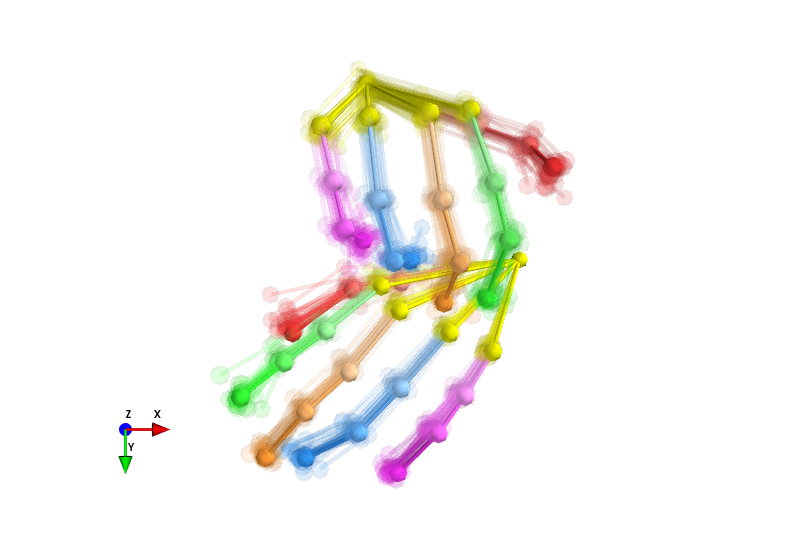}};
  \node[above = 1cm] {\footnotesize{Camera View}};
  \end{tikzpicture}
  \end{subfigure}
  \begin{subfigure}{0.33\columnwidth}
    \begin{tikzpicture}
  \hspace{-0.1cm}\node (image) at (0,0) {\includegraphics[width=\linewidth]{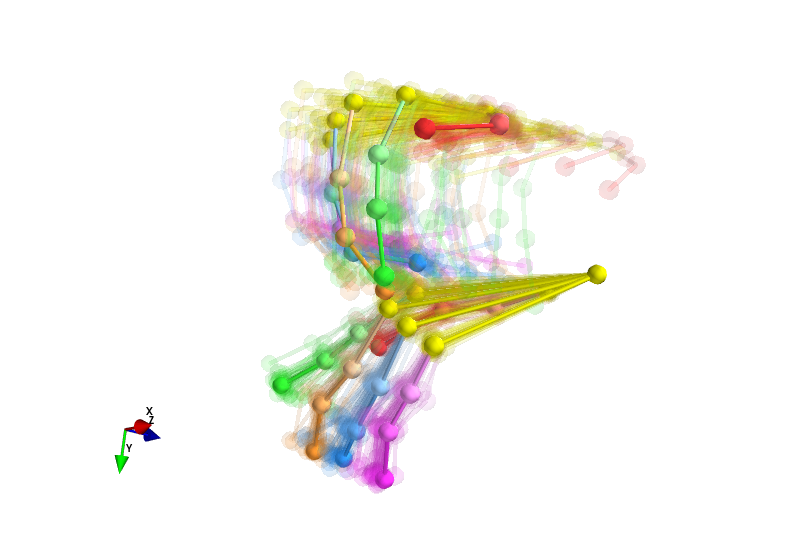}};
  \node[above = 1cm] {\footnotesize{Side View}};
  \end{tikzpicture}
  \end{subfigure}
  \vspace{-.1cm}
  
  \begin{subfigure}{0.33\columnwidth}
  \includegraphics[trim=0 1.2cm 0 1.2cm, clip, width=\linewidth]{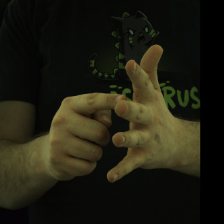}
  \end{subfigure}
  \begin{subfigure}{0.33\columnwidth}
  \includegraphics[trim=2cm 2cm 2cm 1cm, clip, width=\linewidth]{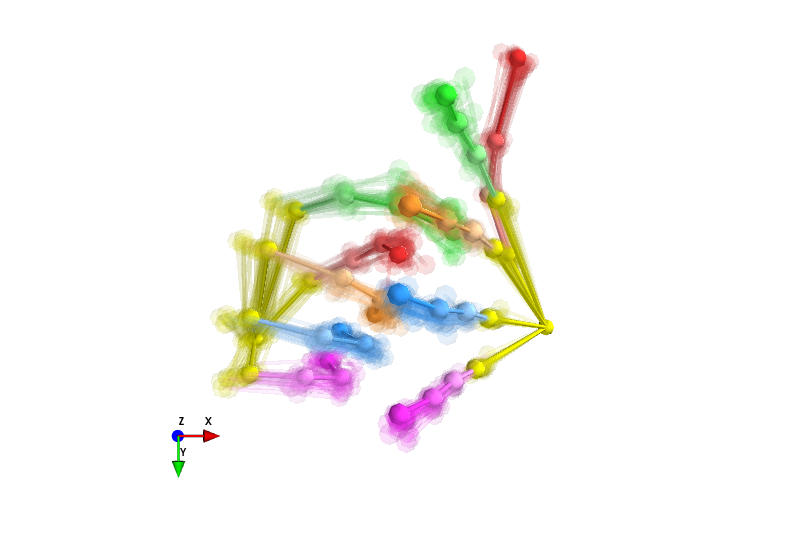}
  \end{subfigure}
  \begin{subfigure}{0.33\columnwidth}
  \includegraphics[width=\linewidth]{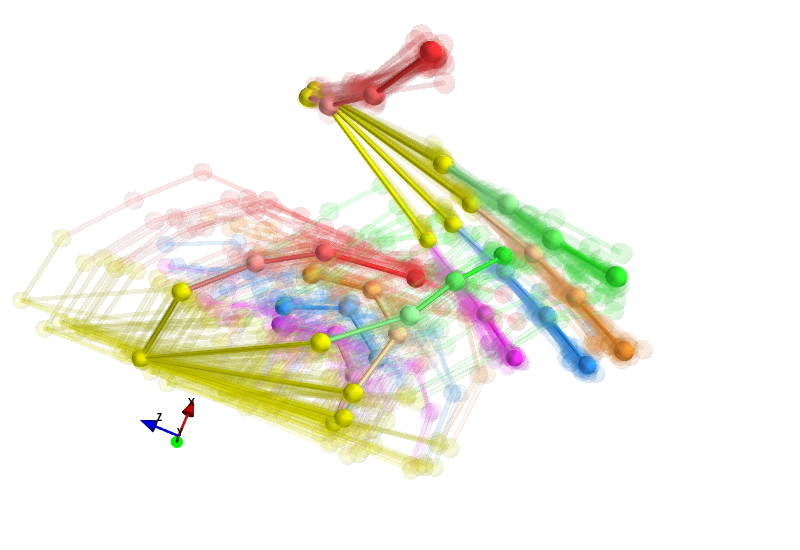}
  \end{subfigure}
  \begin{subfigure}{0.33\columnwidth}
  \includegraphics[trim=0 1.2cm 0 1.2cm, clip, width=\linewidth]{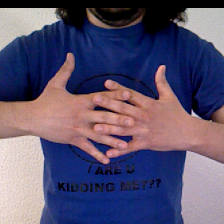}
  \end{subfigure}
  \begin{subfigure}{0.33\columnwidth}
  \includegraphics[width=\linewidth]{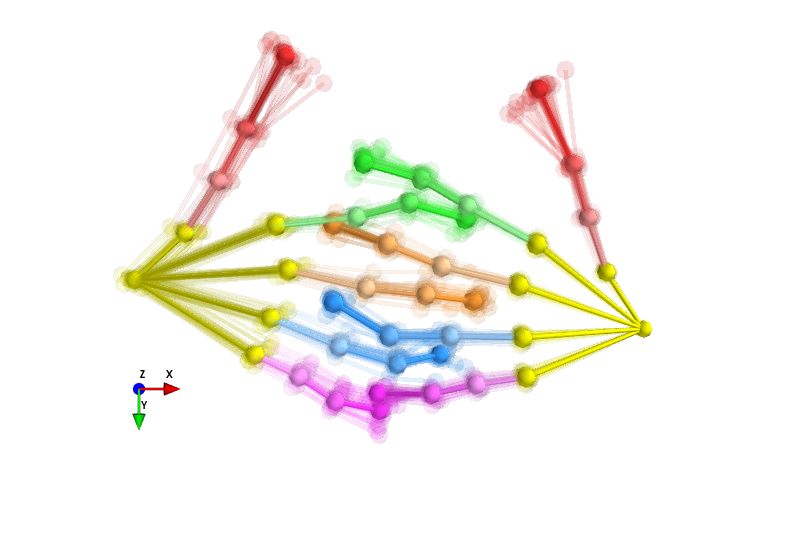}
  \end{subfigure}
  \begin{subfigure}{0.33\columnwidth}
  \includegraphics[width=\linewidth]{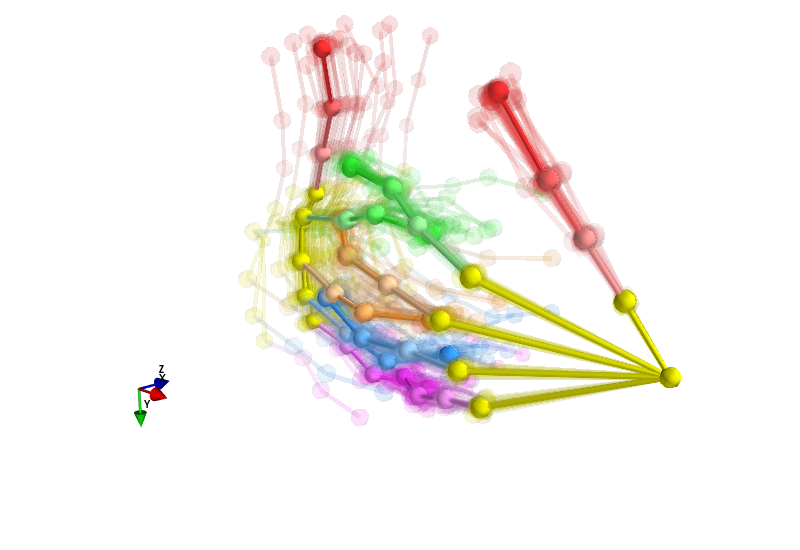}
  \end{subfigure}

  \begin{subfigure}{0.33\columnwidth}
  \includegraphics[trim=0 1.2cm 0 1.2cm, clip, width=\linewidth]{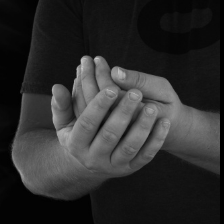}
  \end{subfigure}
  \begin{subfigure}{0.33\columnwidth}
  \includegraphics[trim=2cm 2cm 2cm 1cm, clip, width=\linewidth]{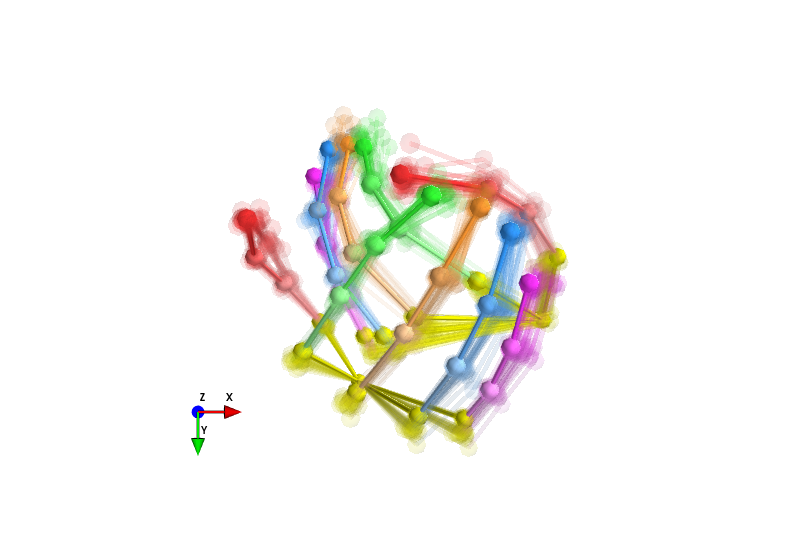}
  \end{subfigure}
  \begin{subfigure}{0.33\columnwidth}
  \includegraphics[trim=2cm 2cm 2cm 1cm, clip, width=\linewidth]{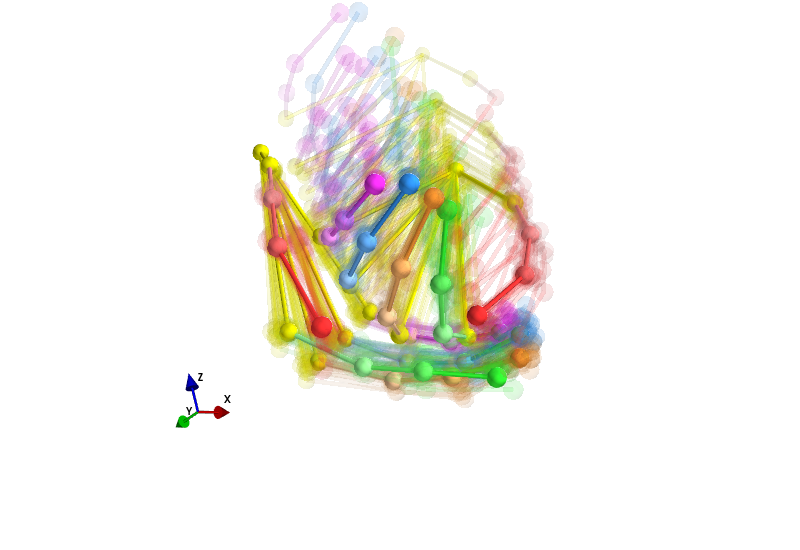}
  \end{subfigure}
  \begin{subfigure}{0.33\columnwidth}
  \includegraphics[trim=0 1.2cm 0 1.2cm, clip, width=\linewidth]{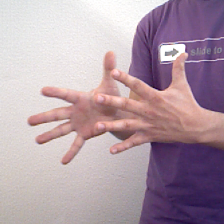}
  \end{subfigure}
  \begin{subfigure}{0.33\columnwidth}
  \includegraphics[width=\linewidth]{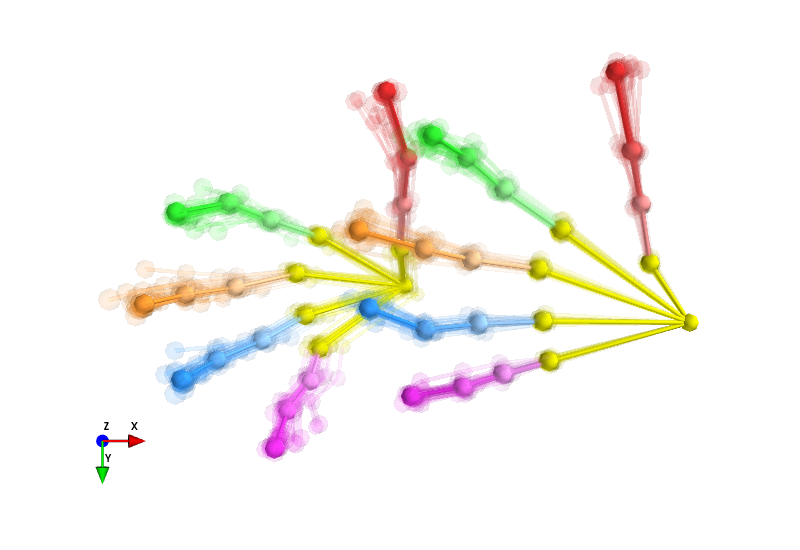}
  \end{subfigure}
  \begin{subfigure}{0.33\columnwidth}
  \includegraphics[width=\linewidth]{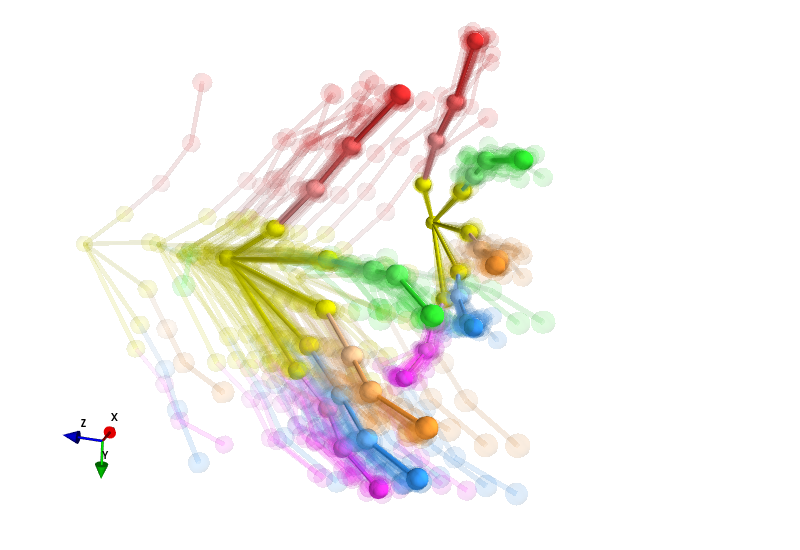}
  \end{subfigure}

  \caption{
  Here we show $30$ pose samples superimposed as semi-transparent skeletons. Samples are aligned to the root joint of one hand and the mode of the distribution is made opaque for ease of visualization. The samples are consistent in the camera view, while showing larger variations in novel views. Examples are from the InterHand2.6M dataset (left) and the Tzionas dataset (right), where we transferred learned 3D ambiguity modeling using only 2D annotations.
  }
  \label{fig:tzionas}
\end{figure*}

\subsection{Ablation Study}
\label{sec:ablation}
\begin{table}[t]
\begin{center}
\resizebox{0.48\textwidth}{!}
{
\begin{tabular}{|l|c|c|c|}
\hline
Method &  Global MMD $\downarrow$ & RRR MMD $\downarrow$  & RR MMD $\downarrow$ \\
\hline\hline
Ours & \textbf{0.50} & \textbf{0.42} & \textbf{0.44}\\
w/o MultiHands & 0.53 & 0.44 & 0.46\\
\hline
w/o $\mathcal{L}_{\text{DetMag}}$ & 0.72 & 0.49 & 0.46\\
w/o $\mathcal{L}_{\mathcal{J}_\text{3D}}$ & 0.65 & 0.62 & 0.52\\
w/o $\mathcal{L}_{\mathcal{J}_\text{2D}}$ & 0.74 & 0.74 & 0.46\\
w/o $\mathcal{L}_{\psi}$ & 0.55 & 0.42 & 0.45\\
w/o $\mathcal{L}_{\theta}$& 0.61 & 0.46 & 0.45\\
\hline
\end{tabular}   
}
\end{center} 

\caption{
All losses and annotations are needed for the best results. 
}
\label{table:comparisonAblation}
\end{table}

We show in Table~\ref{table:comparisonAblation} that every loss helps to make our samples match the ground-truth distribution. 
In particular, our proposed determinant magnitude regularization $\mathcal{L}_{\text{DetMag}}$ is vital for increasing the diversity of 3D samples. 
The mean standard deviation of the joint positions is improved from $18$ to $31$ mm while lowering the MMD.
Lastly, we observe that using multiple annotations from MultiHands in the $\mathcal{L}_{\text{nll}}$ and $\mathcal{L}_{\text{DetMag}}$ terms further improves the MMD, 
which demonstrate the advantage of the differentiable likelihood estimation in the normalizing flow formulation.

\subsection{More Qualitative Results}
\label{sec:qualitative}

In Fig.~\ref{fig:tzionas}, we show qualitative results to demonstrate the diversity and accuracy of our learned pose distribution.
Specifically, we show pose samples visualized as superimposed transparent kinematic skeletons. 
Note that pose variations well reflect the expected monocular ambiguity, and occlusions further increase variability.
Hence, the standard deviation of our samples can serve as an indicator for the ambiguity in the input image and thus uncertainty in the pose prediction.
See supplemental video for more results. 

\subsection{Application: View Selection}
\label{sec:view_selection}
\input{figures/ViewSelectionFigure}

By using the sample standard deviations to estimate pose ambiguity, we can identify camera views that provide the most information for a given motion sequence. 
This information can be useful, for example, in a multi-view capture setup where uninformative cameras can be removed to reduce the hardware and data bandwidth requirements.
We demonstrate this on the InterHand2.6M test set with over $100$ images in the sequence. This consists of the $7$ sequences in Capture0-1 with interacting hands, each with $140$ camera views.

The view quality is evaluated using regret~\cite{berry1985bandit} in MPJPE: the difference between the MPJPE on the selected view and the lowest MPJPE.
The best and worst views selected by our method have a regret of $3.1$ and $15.9$ mm respectively, while the average regret over the cameras is $10.7$ mm. 
This shows that our method is able to eliminate cameras with ambiguous views where the monocular pose estimator is not expected to perform well, while keeping cameras views where the estimator is likely to succeed. 

We can extend view selection to stereo camera pairs by combining two monocular pose distributions.
By assuming conditional independence, we can approximating the pose samples from each view with normal distributions and combine them by taking their product. 
See Fig.~\ref{figure:ViewSelection} and supplemental video for a qualitative evaluation of the selected views.

\section{Limitations and Future Work}
Although we demonstrated promising results, there are some limitations that could be addressed in future work.

Currently, we do not penalize physically implausible intersections in our reconstructions.
As demonstrated in related work~\cite{wang2020rgb2hands,hasson19_obman}, an explicit loss to prevent these intersections could be used to improve the results.

Although we showed promising generalization results on the Tzionas dataset, we did not tackle completely unconstrained in-the-wild images.
We believe that in the future this can be solved with more data, especially 2D annotations for in-the-wild data.

While our experiments verified the need for probabilistic pose estimates in ambiguous scenarios, many applications can only make use of a single pose prediction. 
Future work could investigate ways to integrate additional observations (\eg, temporal information, multi-view images, depth images, task-based priors) to disambiguate the output distribution for a given down-stream task.

\section{Conclusion}
We have presented the first two-hand reconstruction approach to explicitly model the inherent ambiguities that arise from using a single monocular input image.
Given this challenging setting, our method produce a distribution of plausible reconstructions, from which diverse 3D pose samples can be drawn that all explain the observed image evidence. 
Additionally, we showed that existing evaluation schemes are problematic as they assume a single correct pose even though multiple solutions are equally valid. 
Along with our proposed dataset with multiple annotations and the distribution metric, we hope our work demonstrates the need for probabilistic
approaches and provides a way to evaluate them.

\section{Acknowledgments}
The work was supported by the ERC Consolidator Grants 4DRepLy (770784) and TouchDesign (772738) and Spanish Ministry of Science (RTI2018-098694-B-I00
VizLearning). AK acknowledges support via his Emmy Noether Research Group funded by the German Science Foundation (DFG) under Grant No. 468670075.

\bibliographystyle{eg-alpha-doi} 
\bibliography{references}

\end{document}